\pdfoutput=1

\documentclass[11pt]{article}

\usepackage[]{acl}
\usepackage{authblk}
\usepackage{times}
\usepackage{latexsym}
\usepackage{multirow}
\usepackage{multicol}
\usepackage{mathtools}
\usepackage{amsmath,lipsum}

\usepackage[T1]{fontenc}

\usepackage[utf8]{inputenc}

\usepackage{microtype}

\usepackage{inconsolata}

\usepackage{graphicx}

\title{How effective is Multi-source pivoting for \\Translation of Low Resource Indian Languages?}
\author[1]{\bf Pranav Gaikwad}
\author[1]{\bf Meet Doshi}
\author[2]{\bf Raj Dabre}
\author[1]{\bf Pushpak Bhattacharyya}

\affil[1]{CFILT, Indian Institute of Technology Bombay, India}
\affil[2]{NICT, Japan}
\affil[2]{IIT Madras, India}
\affil[1]{\texttt{\{pranavgaikwad,meetdoshi,pb\}@cse.iitb.ac.in}}
\affil[2]{{\texttt{\{prajdabre\}@gmail.com}}}

\begin{document}
\maketitle
\begin{abstract}
Machine Translation (MT) between linguistically dissimilar languages is challenging, especially due to the scarcity of parallel corpora. Prior works \citep{salloum-habash-2013-dialectal,yeshpanov-etal-2024-kazparc-kazakh} suggest that pivoting through a high-resource language can help translation into a related low-resource language. However, existing works tend to discard the source sentence when pivoting. Taking the case of English to Indian language MT, this paper explores the `multi-source translation' approach with pivoting, using both source and pivot sentences to improve translation. We conducted extensive experiments with various multi-source techniques for translating English to Konkani, Manipuri, Sanskrit, and Bodo, using Hindi, Marathi, and Bengali as pivot languages. We find that multi-source pivoting yields marginal improvements over the state-of-the-art, contrary to previous claims, but these improvements can be enhanced with synthetic target language data. We believe multi-source pivoting is a promising direction for Low-resource translation. 
\end{abstract}

\section{Introduction}

The performance of Neural Machine Translation (NMT) models has significantly improved with the Transformer architecture \cite{vaswani2017attention}. However, they require large amounts of data, which is often scarce for many language pairs \cite{DBLP:journals/corr/abs-1907-05019}. In such cases, \textit{Pivoting} is useful, where a source language is translated to an intermediate pivot language, which is then translated into the target language \cite{de2006catalan,utiyama2007comparison}. \citet{toral-etal-2019-neural, yeshpanov-etal-2024-kazparc-kazakh, escolano-etal-2019-bilingual,salloum-habash-2013-dialectal} suggests that pivoting through a related high-resource language (HRL) helps the translation into a low-resource language (LRL). However, our preliminary exploration of English to Indic LRL translation using Indic HRLs as pivots yielded poorer results compared to state-of-the-art systems, likely due to discarding the source language sentence, prompting deeper exploration of pivoting using both source and pivots jointly. 

\citet{zoph-knight-2016-multi} showed that multi-source Neural Machine Translation outperforms one-to-one systems, particularly with distinct sources, guiding us toward multi-source pivoting. We tested various multi-sourcing techniques for English to Indic LRL translation. We applied multiple regularization strategies to balance reliance on source and pivot representations and incorporated various attention schemes. We used two pivots for each language pair to evaluate their impact on translation quality. Finally, we tested the multi-source approach by using multiple pivots simultaneously. 

\textbf{Our contributions are:}
\textbf{a.} A first-of-its-kind study on various multi-source pivot-based machine translation techniques, highlighting the importance of synthetic data for English to Indic pivot-based machine translation. Refer to Table \ref{tab: results} and \ref{tab: tar-syn};
\textbf{b.} Our findings show that multi-source pivoting does not yield significant improvements, even if we use distant \textit{source} and \textit{pivot} languages in case of English to Indic translation, contrary to previous claims;
\textbf{c.} The exploration of the effects of using multiple pivots simultaneously in multi-source machine translation. Refer to Table \ref{tab: results}.


\section{Related Work}




In this section, we discuss works related to pivoting and multi-source NMT.

\citet {utiyama2007comparison, de2006catalan} introduced cascaded pivot-based approach. \citet{zoph-knight-2016-multi, garmash-monz-2016-ensemble} proposed various multi-sourcing approaches. \citet{nishimura-etal-2018-multi,nishimura-etal-2018-multisource, huang-etal-2020-unsupervised-multimodal} devised techniques using incomplete multilingual and synthetic data in multi-sourcing. \citet{libovicky-helcl-2017-attention} investigated various attention mechanisms in multi-source NMT. \citet{machacek-etal-2023-robustness} used multi-sourcing to improve the automatic speech translation systems.

\citet{firat-etal-2016-zero} introduced pivoting with multi-sourcing approaches. \citep{littell-etal-2019-multi} showed that multi-sourcing with a related pivot helps, focusing on Kazak-Russian-English translation.

We focus on translating English to Indic LRLs using a related HRL as a pivot. We study various multi-sourcing techniques and examine the impact of different pivot languages on translation. 

\begin{figure}[htbp]
\centering
  \includegraphics[width=0.42\textwidth]{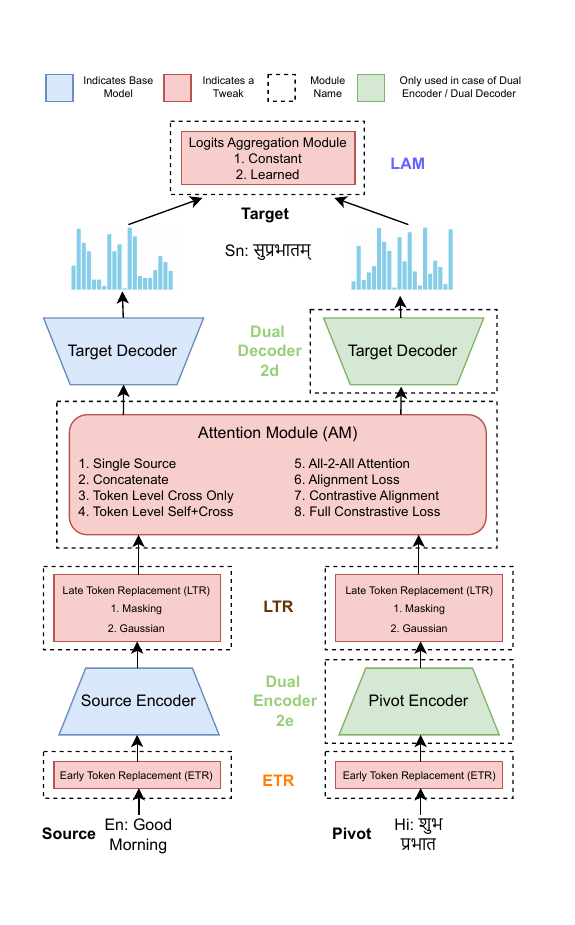}
  \caption{The figure illustrates the various modifications made to the transformer architecture for each experiment, mapping each experiment explained in section \ref{sec: methodology} to the specific architectural changes implemented.}
  \label{fig: arch}
\end{figure}

\section{Methdology}\label{methdology}
\label{sec: methodology}
In this section, we explain the approaches for Multi-source pivoting. The architecture follows Fig. \ref{fig: arch}.

\noindent\textbf{1E-1D:} Our baseline systems include a basic Encoder-Decoder translation model. 

\noindent\textbf{Logits Aggregation Module (LAM):} This method of uses two encoder-decoder pairs, each for source-to-target and pivot-to-target translation. The logits generated by the respective decoders are combined by taking the weighted average \cite{firat-etal-2016-zero}. To predict the $t^{th}$ token $y_t$, each translation path computes the distribution over the target vocabulary, i.e., $P(y_t = w | y_{<t}, X_{src})$ and $P(y_t = w | y_{<t}, X_{pivot})$, which are combined to get the multi-source output distribution:

\begin{math}
P(y_t = w | y_{<t}, X_{src}, X_{pivot}) = 
\alpha * P(y_t = w | y_{<t}, X_{src}) + \beta * P(y_t = w | y_{<t}, X_{pivot}) \hfill (1)
\end{math}

Here, $w$ is a word in a vocabulary, with $X_{src}$ and $X_{pivot}$ being the source and pivot sequence respectively. Initially, we set $\alpha = 0.5$ and $\beta = 0.5$ in our experiments. Subsequently, we explored the possibility of making $\alpha$ and $\beta$ learnable model parameters, by applying a regularisation loss \(\mathcal{L}_{constraint} = (\alpha + \beta - 1)^2\) scaled with $\lambda_{reg}$ to weigh the output distributions.

\begin{table*}[htbp]
\centering
\resizebox{\textwidth}{!}{%
\begin{tabular}{c|cc|cc|cc|cc}
\hline
\multirow{2}{*}{\textbf{Method}} & \multicolumn{2}{c|}{\textbf{En-Ko}}                        & \multicolumn{2}{c|}{\textbf{En-Bo}}                        & \multicolumn{2}{c|}{\textbf{En-Mn}}                        & \multicolumn{2}{c}{\textbf{En-Sn}}                         \\ \cline{2-9} 
                                 & \multicolumn{1}{c|}{\textbf{Pivot-Hi}} & \textbf{Pivot-Mr} & \multicolumn{1}{c|}{\textbf{Pivot-Hi}} & \textbf{Pivot-Bn} & \multicolumn{1}{c|}{\textbf{Pivot-Hi}} & \textbf{Pivot-Bn} & \multicolumn{1}{c|}{\textbf{Pivot-Hi}} & \textbf{Pivot-Mr} \\ \hline
IndicTrans2 (Baseline 1)         & \multicolumn{2}{c|}{\textbf{18.96}}                        & \multicolumn{2}{c|}{16.58}                                 & \multicolumn{2}{c|}{17}                                    & \multicolumn{2}{c}{11.1}                                   \\ \hline
IndicTrans2-FT (Baseline 2)      & \multicolumn{2}{c|}{16.12}                                 & \multicolumn{2}{c|}{\textit{17.45}}                        & \multicolumn{2}{c|}{\textit{20.34}}                        & \multicolumn{2}{c}{\textbf{12.82}}                         \\ \hline
Cascade pivot (Baseline 3)       & \multicolumn{1}{c|}{18}                & 18                & \multicolumn{1}{c|}{16.9}              & 16.1              & \multicolumn{1}{c|}{15.9}              & 15.5              & \multicolumn{1}{c|}{10.7}              & 10.7              \\ \hline
2E-1D                            & \multicolumn{1}{c|}{17.15}             & 16.83             & \multicolumn{1}{c|}{17.01}             & 17.26             & \multicolumn{1}{c|}{20.55}             & 20.45             & \multicolumn{1}{c|}{12.48}             & 12.44             \\ \hline
2E-2D + LAM 1                    & \multicolumn{1}{c|}{16.45}             & 16.83             & \multicolumn{1}{c|}{16.48}             & 16.07             & \multicolumn{1}{c|}{18.77}             & 18.52             & \multicolumn{1}{c|}{12.19}             & 12.06             \\ \hline
2E-2D + LAM 2                    & \multicolumn{1}{c|}{16.53}             & 17.11             & \multicolumn{1}{c|}{16.45}             & 16.86             & \multicolumn{1}{c|}{19.79}             & 20.13             & \multicolumn{1}{c|}{12.31}             & \textit{12.66}    \\ \hline
2E-1D + ETR                      & \multicolumn{1}{c|}{16.31}             & 16.5              & \multicolumn{1}{c|}{15.77}             & 17.07             & \multicolumn{1}{c|}{19.5}              & 20.26             & \multicolumn{1}{c|}{12.21}             & 12.44             \\ \hline
2E-1D + LTR 1                    & \multicolumn{1}{c|}{16.83}             & 16.86             & \multicolumn{1}{c|}{16.52}             & 16.89             & \multicolumn{1}{c|}{19.81}             & 19.97             & \multicolumn{1}{c|}{11.66}             & 12.08             \\ \hline
2E-1D + LTR 2                    & \multicolumn{1}{c|}{16.7}              & \textit{17.45}    & \multicolumn{1}{c|}{16.59}             & 16.86             & \multicolumn{1}{c|}{20.19}             & 20.36             & \multicolumn{1}{c|}{11.64}             & 12.21             \\ \hline
2E-1D + AM 4                     & \multicolumn{1}{c|}{16.85}             & 17.11             & \multicolumn{1}{c|}{17.02}             & \textbf{18.09}    & \multicolumn{1}{c|}{20.85}             & 20.66             & \multicolumn{1}{c|}{12.42}             & 12.65             \\ \hline
2E-1D + AM 6                     & \multicolumn{1}{c|}{16.82}             & 17.1              & \multicolumn{1}{c|}{17.57}             & 17.03             & \multicolumn{1}{c|}{20.76}             & 20.82             & \multicolumn{1}{c|}{12.46}             & 12.28             \\ \hline
2E-1D + AM 7                     & \multicolumn{1}{c|}{16.89}             & 16.87             & \multicolumn{1}{c|}{17.4}              & 17.13             & \multicolumn{1}{c|}{20.61}             & 20.63             & \multicolumn{1}{c|}{12.07}             & 12.23             \\ \hline
2E-1D + AM 8                     & \multicolumn{1}{c|}{16.82}             & 16.31             & \multicolumn{1}{c|}{17.35}             & 17.29             & \multicolumn{1}{c|}{\textbf{20.86}}    & 20.6              & \multicolumn{1}{c|}{12.55}             & 12.5              \\ \hline
MS-MP-MT                       & \multicolumn{2}{c|}{4.68}                                  & \multicolumn{2}{c|}{5.06}                                  & \multicolumn{2}{c|}{6.38}                                  & \multicolumn{2}{c}{3.6}                                    \\ \hline
\end{tabular}
}
\caption{Results on Pivot-Synthetic Multi-source pivoting.`En-X' represents English to X translation and Pivot-Y represents pivot Y, where X instantiates to En: English, Ko: Konkani, Bo: Bodo, Mn: Manipuri, and Sn: Sanskrit and Y to Hi: Hindi, Mr: Marathi, and Bn: Bengali. Each experiment is mapped with architectural changes explained in Section \ref{sec: methodology} and Fig. \ref{fig: arch}}
\label{tab: results}
\end{table*}

\noindent\textbf{Attention Module (AM):} This combines representations from multiple encoders and produces aligned representations. Variations of the AM are:

\textbf{1) Single Source:} This is an identity function that passes the source representation to the decoder.

\textbf{2) Concatenation:} The representations of both encoders are concatenated and fed to the decoder. 

\textbf{3) Token Attention Cross Only:} After the attention module receives contextualised representations, $x_1^{s}, x_2^{s}, ..., x_m^{s}$ from the \textit{src} encoder and $x_1^{p}, x_2^{p}, ..., x_n^{p}$ from the \textit{pivot} encoder, this module takes a \textit{src} representations $x_{\{1,...,m\}}^s$ and calculates cross attention with each token from \textit{pivot} representations $x_{\{1,...,n\}}^p$ and vice versa. These token-level features are concatenated with each other and then fed to the decoder. 

\textbf{4) Token Attention Cross Concat:} We concatenate the ``Token Attention Cross only'' embeddings with self-attention generated by the encoders. 

\textbf{5) All-2-All Attention:} In All-2-All attention we concatenate both the representations by the encoders and then calculate self-attention over a sequence \{$x_1^{s}, x_2^{s}, ..., x_m^{s}, x_1^{p}, x_2^{p}, ..., x_n^{p}$\}.

\textbf{6) Alignment Loss:} We include cosine embedding loss between the \textit{src} and \textit{pivot} sentence representations of the encoder for alignment.

\textbf{7) Contrastive Alignment:} Following \citet{moiseev-etal-2023-samtone}, we include a contrastive loss with in-batch negatives.

\textbf{8) Full Contrastive Alignment:} We include a three-way contrastive loss between \textit{src}, \textit{pivot} encoder and the \textit{tgt} decoder representations.

\noindent\textbf{Token Replacement Regularization:} We explore the following dropout regularization approaches: 

\textbf{1) Early Token Replacement (ETR):} We replace $p$\% of the token embeddings from both source and pivot encoders at random with a \texttt{[MASK]} token embedding. 

\textbf{2) Late Token Replacement (LTR):} We drop $p$\% of the embeddings from each encoder. The remaining embeddings are concatenated and fed to the target decoder. We explore two strategies: learned masking with a \texttt{[MASK]} token, and Gaussian masking by adding Gaussian noise to embeddings. 

\noindent\textbf{2E-2D:} This system consists of two 1E-1D models stitched together using a Logits Aggregation Module (LAM), where each 1E-1D model performs \textit{src-to-tgt} and \textit{pivot-to-tgt} translation. We try two variations of the LAM module: 1) Uniform weighing and 2) Learned Weighing.

\noindent\textbf{2E-1D:}  Optimizing two decoders at once is difficult \cite{le-etal-2020-dual}, we drop one of the decoders in the 2E-2D architecture that was trained along with the pivot and instead merge the output of both the encoders to feed into a single decoder using an Attention Module (AM). 

\noindent\textbf{Multi Source Multi Pivot Machine Translation (MS-MP-MT):}\label{msmpmt} In this approach, we use multiple pivots along with source in the Multi-sourcing framework. The final probability for $t^{th}$ token $y_t$ is

\(\{P(y_t=w\mid y_{<t},X_{src},Z_{pivot}^1,...,Z_{pivot}^k)\)
\(= \sum_{i=1}^{k} \left\{ P(Y \mid X, Z_{pivot}^i) \cdot P(Z_{pivot}^i\mid X) \right\}\) \hfill (2)
Where $w$ is a vocabulary word, $X$ is the source language sentence and $Z_{pivot}^1$ to $Z_{pivot}^k$ are pivot language representations. $P(Y \mid X_{src}, Z_{pivot}^i)$ is extracted through the logits generated by the decoder similar to the working of a 2E-1D model. $P(Z_{pivot}^i\mid X_{src})$ is calculated using frozen scorer model (IndicTrans2). 


\section{Experiments and Results}\label{sec:experiments}
In this section, we describe the training procedure, datasets, and results for the models in Sec. \ref{methdology}.

\subsection{Data}
We utilize Bharat Parallel Corpus Collection BPCC for our experiments. We utilize two types of data for our experiments:
\noindent\textbf{Pivot-synthetic:} We take \textit{src-tgt} original data and translate the \textit{src} into the pivot language. 
\noindent\textbf{Target-synthetic:} We use \textit{src-pivot} parallel data and translate it into \textit{tgt}. More details of the size of training data used can be found in Table \ref{tab: data} and Appendix \ref{app: data statistics}. We evaluate the multi-source pivoting techniques discussed in Section \ref{methdology} for English-to-X translations, where `X' is Konkani, Bodo, Manipuri, and Sanskrit. We used Hindi as the pivot for all language pairs, Marathi for Konkani and Sanskrit, and Bengali for Bodo and Manipuri.

\begin{table}[htbp]
\centering
\resizebox{0.85\columnwidth}{!}{%
\begin{tabular}{c|c|c}
\hline
 \textbf{Method} & \textbf{En-Ko}      & \textbf{En-Bo}                     \\ \hline
IndicTrans2 (Baseline 1)         & \textit{18.96}      & 16.58                               \\ \hline
IndicTrans2-FT (Baseline 2)      & 18.67               & \textit{17.5}                       \\ \hline
Cascade pivot (Baseline 3)       & 18                  & 16.9                                \\ \hline
2E-1D + AM 1                     & 19.14               & 17.5                                \\ \hline
2E-2D + LAM 1                    & 18.6                & 16.42                               \\ \hline
2E-2D + LAM 2                    & 18.99               & 17.48                               \\ \hline
2E-1D + ETR                      & 19.13               & 17.64                               \\ \hline
2E-1D + LTR 1                    & 18.84               & 17.34                               \\ \hline
2E-1D + LTR 2                    & 19.08               & 17.42                               \\ \hline
2E-1D + AM 3                     & 18.84               & 17.56                               \\ \hline
2E-1D + AM 4                     & 19.15               & 17.76                               \\ \hline
2E-1D + AM 5                     & 19.12               & 17.4                                \\ \hline
2E-1D + AM 5 + AM 6              & \textbf{19.18}      & \textbf{18.19}                      \\ \hline
2E-1D + AM 6                     & 19.17               & 17.56                               \\ \hline
2E-1D + AM 7                     & 19.04               & 17.7                                \\ \hline
2E-1D + AM 8                     & 19.06               & 17.57                               \\ \hline
\end{tabular}}
\caption{Results on Target-Synthetic Multi-source pivoting. Where En-Ko is English to Konkani and En-Bo is English to Bodo translation. Each experiment is mapped with architectural changes explained in Section \ref{sec: methodology}} 
\label{tab: tar-syn}
\end{table}

\subsection{Model and Training}
We use IndicTransV2-1.1B \cite{ai4bharat2023indictrans2} as our base model for the initialization of encoder and decoder models. For pivot-synthetic we use the number of steps proportional to the size of the \textit{src-tgt} parallel data and for target-synthetic we limit ourselves to $20k$ steps. We use a learning rate of $3e^{-5}$ with a batch size of 64. More hyperparameters can be found in Appendix \ref{app: hyperparameters}.

\subsection{Results and Analysis}
\label{sec: results analysis}

\noindent\textbf{Baselines:} We used IndicTransV2 \cite{ai4bharat2023indictrans2} as our primary baseline, which is state-of-the-art for English-to-Indic translation. For each language pair, fine-tuned IndicTransV2 is used as another baseline. We implemented cascaded pivoting with IndicTransV2, but it failed to outperform the state-of-the-art.

\noindent\textbf{Does pivot-based Multi-sourcing help?} We began our experiments with the basic multi-source logits aggregation (LAM) technique. Initially, the 2E-2D logits averaging method performed poorly. However, when we applied learned weights to this method, we observed a slight improvement. After training, the weight distribution for source and pivot languages was almost uniform, except for the English-to-Sanskrit pair where English had a higher weight, as illustrated in Figure \ref{fig: weighing}. Next, we explored the 2E-1D approach, which outperformed the ``2E-2D logits averaging" method for most language pairs but still underperformed compared to baseline methods. Example translations are provided in Appendix \ref{QA}.

\noindent\textbf{Does regularization help?} We tested dropout regularization to enhance training robustness by applying ``Early Token Replacement," ``Late Token Replacement," and ``Gaussian noise addition" to 20\% of encoder representations. The results of these experiments were mixed. To further investigate the effects of regularization, we conducted a study detailed in Appendix \ref{app:learned weighing}), which showed that increasing the amount of noise can only give slight improvements.

\noindent\textbf{Increasing alignment:} We utilize different types of attention mechanisms to improve language alignment between \textit{src} and \textit{pivot} languages. The concatenating cross-attention with self-attention approach performs slightly better than most previously discussed methods. We also experimented with alignment loss that used cosine similarity to calculate alignment producing mixed results. 

\noindent\textbf{Do linguistically distant sources help?} \cite{zoph-knight-2016-multi} claims that using linguistically distant sources helps in multi-source translation. We utilize Indian HRLs as pivots, linguistically distant from the source English, for English to Indic LRL translation. We do observe minor improvements with linguistically distant sources. In contrast to \citet{zoph-knight-2016-multi}, our findings show that for 3 out of 4 \textit{src-tgt} language pairs, results are on average better when the pivot is from the same language family as the target.

\noindent\textbf{Does more data help?} We exploited the pivot language by using source-pivot parallel data to create a synthetic multi-way parallel corpus. This leads to improvements as compared to baselines (Table. \ref{tab: tar-syn}) for English to Konkani and English to Bodo translation but this cannot be considered statistically significant and can be due to a distillation effect. 

Our extensive experimentation with multi-source NMT showed only minor improvements, even when using distant languages as source and pivot. Contrary to \citet{zoph-knight-2016-multi}, our findings indicate that multi-source approaches offer marginal gains over baseline methods for low-resource Indic languages, warranting further exploration.

\section{Conclusion and Future work}
In this paper, we explore pivot-based machine translation for English to low-resource Indic languages using multi-source techniques. Through extensive experimentation with various methods and pivot languages, we demonstrate that distant sources and pivots do not necessarily improve results. Our findings indicate that there is no one-size-fits-all approach, but the right combination of methods and pivots enhances translation quality. We also observe that data quantity, including synthetic data, significantly influences outcomes. We conclude that multi-source pivoting is a promising direction warranting further investigation.

\section*{Limitations}
\begin{enumerate}
    \item For pivot-synthetic experiments we are limited by the size of the source-target parallel corpus available. 
    \item For target-synthetic experiments we limit ourselves to 1 Million parallel sentences because our aim was to compare various multi-sourcing experiments. 
\end{enumerate}
\bibliography{custom}

\appendix

\section{Training Data}
\label{app: data statistics}

\subsection{Data}
We use the Bharat Parallel Corpus Collection (BPCC) which is a collection of English-Indic parallel corpora for 22 scheduled Indian languages, as the training set \cite{ai4bharat2023indictrans2}. For every target language `X', we utilize the `En-X' parallel corpus from BPCC and create multi-way parallel data synthetically, by translating the English side to the chosen pivot languages. The dataset details are given in table \ref{tab: data}. We also synthetically create a 4-way parallel corpus comprising 1 Million sentences each for two target languages namely Bodo and Konkani. We use the English-Hindi parallel corpus and translate the English side to Bodo and Konkani, and the respective pivot langues chosen for them. We use \citet{ai4bharat2023indictrans2} model for all synthetic generations. We evaluate all of our experiments on the IN22-Conv and IN22-Gen test set which is a benchmark for evaluating machine translation performance created by \citet{ai4bharat2023indictrans2}.

\subsection{Preprocessing}
 We use IndicTransV2 as our base model so we follow the same preprocessing steps as \citet{ai4bharat2023indictrans2}. We apply standard preprocessing, which includes removing redundant spaces, removing special characters, and normalizing punctuations. Additionally, we convert the Indic numerals to English numerals using a dictionary-based mapping. We apply rule-based script conversion using the IndicNLP library \cite{kunchukuttan2020indicnlp} to represent most of these languages in a single script (Devanagari). Thus, effectively our models are trained with three scripts for Indic languages Meitei (Manipuri), Latin (English), and Devanagari (for all remaining Indic languages). We use two separate tokenizers based on the byte-pair-encoding (BPE) algorithm \citep{sennrich-etal-2016-neural} using the SentencePiece library \cite{kudo-richardson-2018-sentencepiece} for English and Indic languages. The vocabulary size is 32K and 128K for English and Indic SPM models, respectively. After tokenization, the source language and target language tags are prepended to all the sentences.

\begin{table}[htbp]
\centering
\begin{tabular}{|c|c|}
\hline
Language pair     & \#Parallel sentences \\ \hline
English-Konkani   & 98k                  \\ \hline
English-Bodo      & 117k                 \\ \hline
English-Manipuri & 43k                 \\ \hline
English-Sanskrit & 278k                 \\ \hline
\end{tabular}
\caption{The table shows dataset details, where \#Parallel sentences is the number of parallel sentences present in the corpus.}
\label{tab: data}
\end{table}

\begin{table}[htbp]
\centering
\begin{tabular}{|c|c|c|}
\hline
Language pair     & \#Epochs & \#Save steps \\ \hline
English-Konkani   & 4    & 2000                  \\ \hline
English-Bodo      & 4      & 2000           \\ \hline
English-Manipuri & 6        & 1250        \\ \hline
English-Sanskrit & 4         & 5000       \\ \hline
\end{tabular}
\caption{The table shows the training details for each language pair, where \#Epochs is the number of training epochs and \#Save steps is the number of steps after which the checkpoints are saved.}
\label{tab: training}
\end{table}

\section{Training Details}
\label{app: hyperparameters}

\subsection{Model and Training}
We use IndicTransV2 as our base model so we follow the same preprocessing steps as \citet{ai4bharat2023indictrans2}. It is a multilingual model based on transformer architecture. The architecture comprises 18 encoder layers and 18 decoder layers, an input dimension of 1024, pre-normalization \cite{xiong2020layer} for all modules, a feedforward dimension of 8192, and 16 attention heads. The total parameter count is 1.1B. We use the GELU activation \cite{hendrycks2016gaussian} and the Inverse-square root learning rate scheduler with 1000 warmup updates. We use the batch size of 48 for all our models. The number of training steps and the interval after which the checkpoints are stored is shown in table \ref{tab: training}. We decide these hyperparameters based on the amount of available parallel data for training. We have used the HuggingFace library for the training of all our models. We have used the PyTorch library to implement ensemble learning. We evaluate the translation quality of our models by BLEU \citep{papineni-etal-2002-bleu}. We have used Nvidia-DGX-H100 GPUs with 80 GB memory for data synthetically, by translating the English side to models using multiple encoder-decoder pairs and Nvidia-DGX-A100 GPUs with 40 GB memory for models with multiple encoders and single decoder. 
\section{Qualitative Analysis}\label{QA}
We show the quality of the translation, compared to IndicTransV2 and reference translation. We provide translation and gloss for each non-English example in English. 
\begin{figure*}[]
    \centering
    \includegraphics[width=0.99\textwidth]{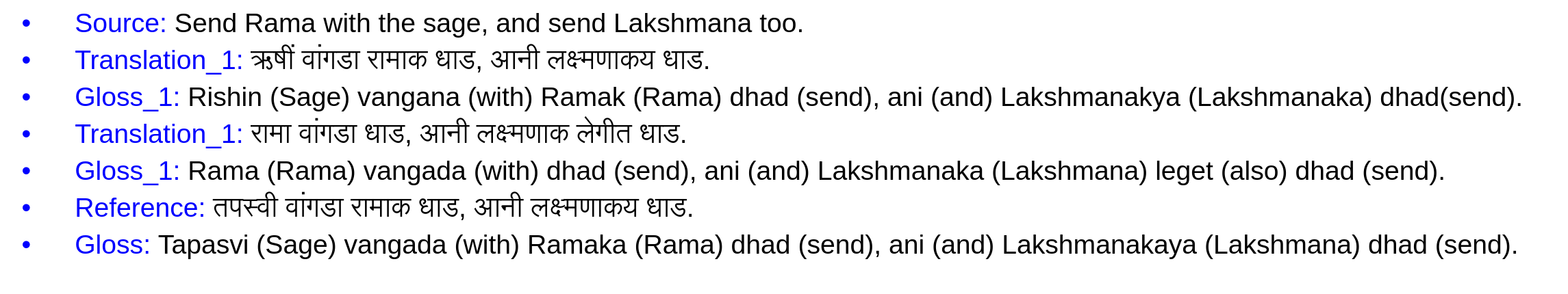}
    \caption{This is an example translation from English to Konkani. Where Translaion\_1 and Gloss\_1 represent the translation produced by IndicTransV2, Translaion\_2, and Gloss\_2 represent the translation produced by our system (2E-1D). Reference and Gloss belong to the reference sentence.}
    \label{fig: qualitative1}
\end{figure*}
\begin{figure*}[]
    \centering
    \includegraphics[width=0.99\textwidth]{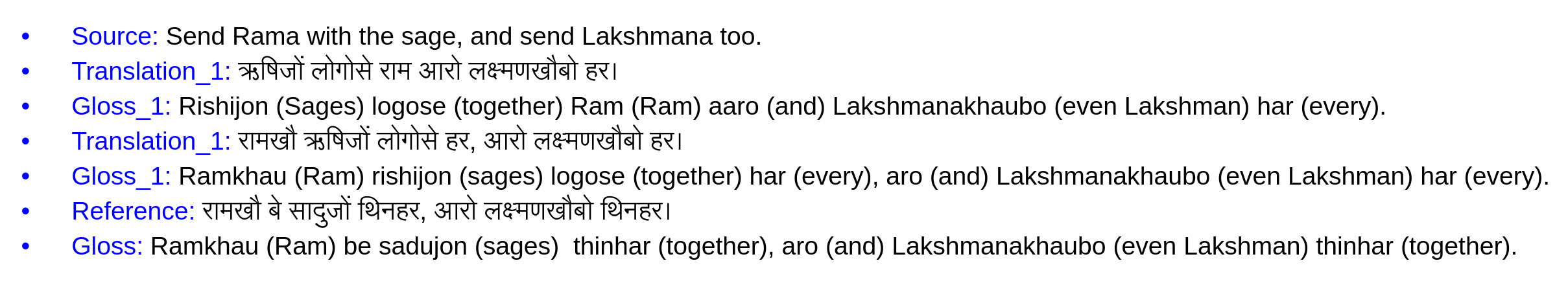}
    \caption{This is an example translation from English to Bodo. Where Translaion\_1 and Gloss\_1 represent the translation produced by IndicTransV2, Translaion\_2, and Gloss\_2 represent the translation produced by our system (2E-1D). Reference and Gloss belong to the reference sentence.}
    \label{fig: qualitative2}
\end{figure*}
\section{Additional Results}

\begin{figure}[]
    \centering
    \includegraphics[width=0.4\textwidth]{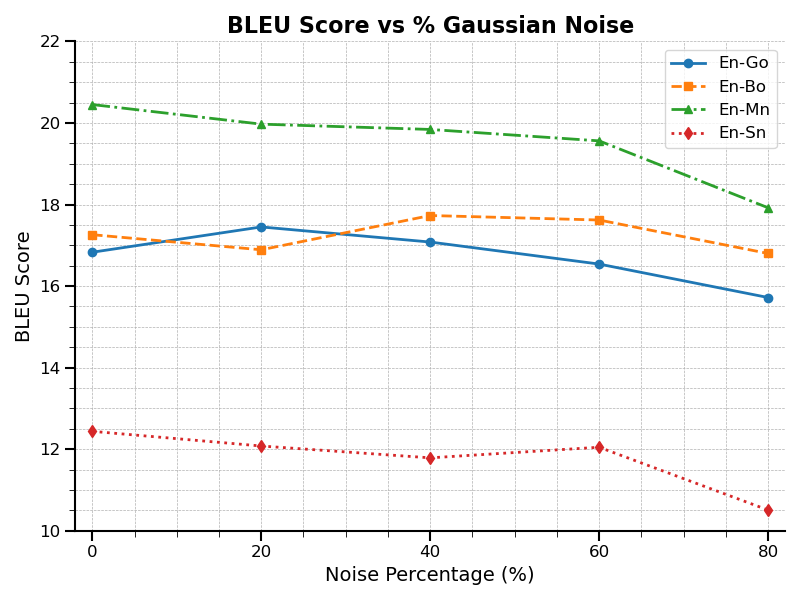}
    \caption{Effect of increasing Gaussian noise during training}
    \label{fig:ablation gaussian}
\end{figure}

\subsection{Dropout regularization for Multi-sourcing}
In this section, we discuss various dropout regularization approaches that we explored in multi-source scenarios. 

\subsubsection{Early token dropout}
In this approach, we perform random masking of tokens on the source and pivot sides. We replace  $p$\% of the token embeddings from the source and pivot encoders with a mask token embedding. This prevents the model from being completely reliant on one of the encoders for all information.

\subsubsection{Late token dropout}
In this approach, we pass the complete source and pivot sentence to the source and pivot encoder respectively. After the contextual embeddings are generated by the encoders, $p$\% of the embeddings each from the source and pivot encoders are dropped. Then the source and pivot encoder embeddings are concatenated and fed to the target decoder. 

\subsubsection{Noise addition}
In this approach, $p$\% of token embeddings each from the source and pivot side are randomly selected and corrupted by adding Gaussian noise. The model is forced to attend to the pivot language representation of a token if the source language representation is corrupted due to the addition of noise and vice versa. We perform experiments by gradually increasing the amount of to observe the effect on translation quality. Figure \ref{fig:ablation gaussian}
shows the effect of increasing gradual noise added to the \textit{src} and \textit{pivot} representations. Even after adding 80\% of noise, we only see a drop of 2 BLEU points across languages, suggesting that the model is forced to carry more information across token representations.
\label{app: additional results}

\section{Learned Weighing}
\label{app:learned weighing}

\begin{figure}[htbp]
\centering
  \includegraphics[width=0.45\textwidth]{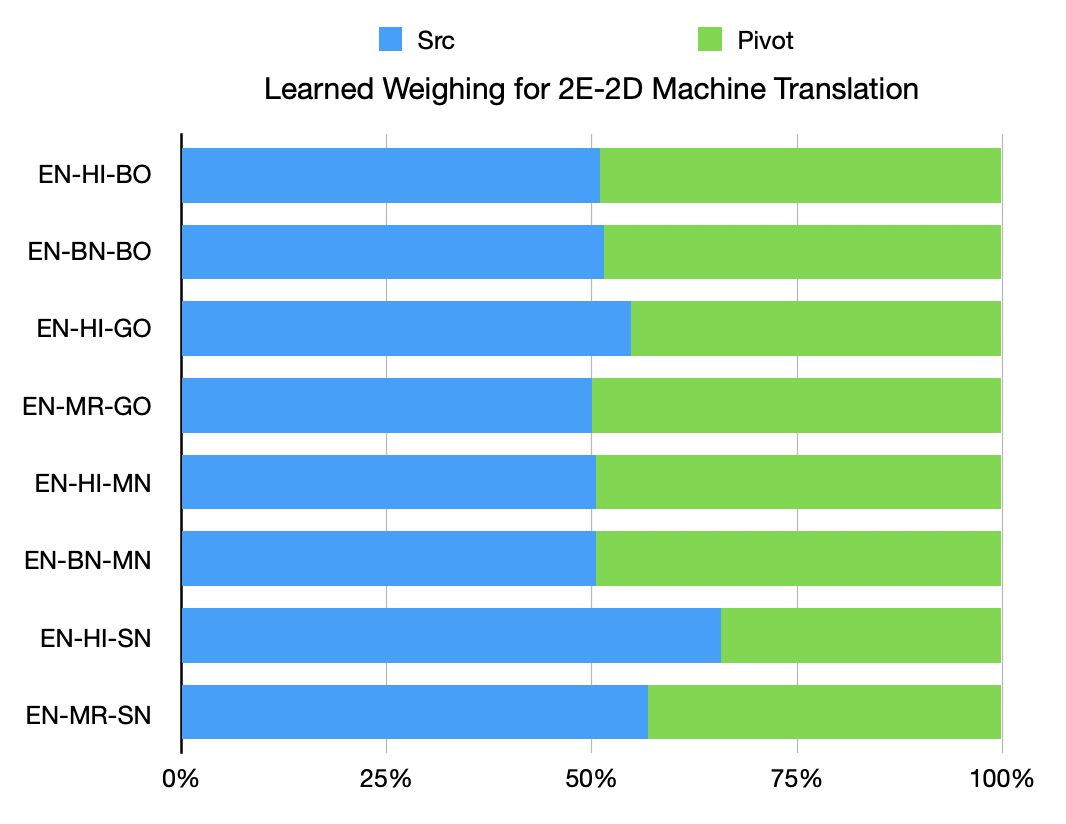}
  \caption{The figure illustrates the weights learned by the Logits Aggregation Module (LAM) after training. For most of the cases, we see an equal weightage given to both source and pivot side logits except in the case of Hindi as a pivot for Sanskrit.}
  \label{fig: weighing}
\end{figure}

\end{document}